\documentclass{article}

\usepackage{arxiv}
\usepackage{amsmath}
\usepackage[utf8]{inputenc} 
\usepackage[T1]{fontenc}    
\usepackage{hyperref}       
\usepackage{url}            
\usepackage{booktabs}       
\usepackage{amsfonts}       
\usepackage{nicefrac}       
\usepackage{microtype}      
\usepackage{lipsum}		
\usepackage{graphicx}
\usepackage{natbib}
\usepackage{doi}
\usepackage{xcolor}

\usepackage{tcolorbox}
\title{Frequency-Conditioned Privacy Auditing of Fine-Tuned Language Models}

\author{ 
{Aymen Noor}\\
	Department of Computer Science\\
	Miami University\\
	  Ohio \\
	\texttt{noora@miamioh.edu} \\
	\And
    {Ke Lyu} \\
	Department of Computer Science\\
	Miami University\\
	Ohio\\
	\texttt{lyuk4@miamioh.edu} \\
    \And
    {Honglu Jiang} \\
	Department of Computer Science\\
	Miami University\\
	Ohio\\
	\texttt{jiangh34@miamioh.edu} \\
}

\renewcommand{\shorttitle}{\textit{arXiv} Template}

\hypersetup{
pdftitle={A template for the arxiv style},
pdfsubject={q-bio.NC, q-bio.QM},
pdfauthor={Aymen Noor, Honglu Jiang},
pdfkeywords={First keyword, Second keyword, More},
}

\begin{document}
\maketitle

\begin{abstract}
Large language models are increasingly fine-tuned on sensitive data, where the same person's information may appear multiple times across records. This can cause models to memorize sensitive information, but memorization does not necessarily mean that the information can be recovered. A more serious privacy risk arises when the model begins to reproduce that information during generation. In this paper, we study how often sensitive information must appear in the fine-tuning data before it becomes recoverable. Across four language models, we vary the frequency of email addresses and phone numbers from $1\times$ to $50\times$ and compare standard fine-tuning, gradient clipping, and
differentially private training. We find that the number of repetitions required before sensitive information becomes recoverable varies substantially across models. Gradient clipping can delay extraction, but it does not consistently prevent leakage at high repetition. DP-SGD at $\varepsilon=5$ produces zero observed extraction through $50\times$ repetition across all four models under the attacks we test, although the utility cost varies by model. These findings characterize how observed extraction varies across sensitive-data frequencies, model families, and training mechanisms, and highlight substantial differences in extraction behavior across
models. Code and data are publicly available.\footnote{
Code: \url{https://github.com/MN-NR/dp-llm}.
Data: \url{https://huggingface.co/datasets/mnoor/Dp_llm}.
}
    
\end{abstract}


\section{Introduction}
Large language models are increasingly fine-tuned on data from domains where privacy matters, including healthcare, finance, customer support, and internal organizational data. In such datasets, the same person's information may appear multiple times across different records. Repeated information is more likely to be memorized and regenerated by language
models, creating a risk that sensitive training data may later be revealed during generation \citep{carlini2021extracting,kandpal2022deduplicating,carlini2022quantifying,mireshghallah2022empirical}.

However, statistical evidence of memorization does not necessarily imply practical extractability. A loss difference or membership-inference score may distinguish training examples from unseen examples without revealing whether the underlying sensitive information can actually be recovered through generation.
The more serious privacy failure occurs when information that was present during training becomes recoverable from the model. Recent work has increasingly highlighted this difference between detecting that an example influenced a model and recovering the information itself
\citep{maricato2026leakitperdocumentextraction,
liu2026beyond, che2026memorization}. Prior work has largely studied memorization and extraction as separate privacy risks, evaluating the former through loss- or membership-based signals \citep{mireshghallah2022empirical,chen2026window}
and the latter through generation-based attacks such as canary, prefix-conditioned, and targeted extraction \citep{carlini2019secret,carlini2021extracting,hayes2025measuring}. 
We study the transition from memorization to extraction during downstream fine-tuning and raise a practical question: \emph{how does observed extraction change as sensitive information appears more frequently during fine-tuning?}
Answering this question requires more than showing that repetition is
associated with leakage. Naturally occurring frequency can be mixed with
many other properties of the data, and evaluating extraction at only one
or two repetition levels does not reveal where the transition begins.
Moreover, different models may not reach this point at the same frequency,
and privacy-preserving training may change the transition in different
ways. We therefore ask three questions: (1) how does extraction change as
the frequency of sensitive information increases, (2) does the amount of
repetition required for recovery differ across models, and (3) how do
gradient clipping and DP-SGD change this behavior and the utility of the
model? We also ask whether commonly used privacy scores can identify which
individual examples will actually become extractable.

We study these questions in a controlled fine-tuning setting across four
language models. We vary the frequency of email addresses and phone
numbers from $1\times$ to $50\times$ and measure both statistical
memorization and direct extraction. We compare matched non-private,
clipping-only, and DP-SGD training so that the effect of gradient clipping
can be separated from the additional effect of DP noise. We further test
whether privacy scores that distinguish training members from unseen
examples can also identify which sensitive values become recoverable
through generation.

Our results show that the number of repetitions required before sensitive
information becomes recoverable differs substantially across models.
Gradient clipping can delay extraction, but its effect is not uniform
across models or repetition levels. In contrast, DP-SGD at
$\varepsilon=5$ yields zero observed extraction through $50\times$
repetition across all four models. This holds under both greedy and stochastic attacks, although the utility cost varies by model. We
also find that strong statistical privacy signals do not necessarily
identify the individual examples that will later be extracted.

\section{Related Work}
\textbf{2.1 Repetition, Memorization, and Extraction}

Language models can memorize and reproduce examples from their training data, creating a risk that sensitive information may be recovered through
generation \citep{carlini2019secret,carlini2021extracting}. Repetition is
one of the strongest factors affecting this behavior. Duplicate training
sequences are regenerated much more frequently, and memorization increases
with duplication count, model capacity, and the amount of context provided
at inference time
\citep{kandpal2022deduplicating,carlini2022quantifying}. These findings motivated data deduplication as one way to reduce training-data leakage. Kandpal et al., for example, find a strong relationship between a sequence's training count and its regeneration rate.

Extraction itself can also depend on how the model is queried. A training
example that is not recovered by greedy decoding may still be recoverable
through repeated sampling, motivating probabilistic rather than purely
binary views of extraction \citep{hayes2025measuring}. Recent work
therefore provides strong evidence that repetition affects memorization
and extraction, but most studies either analyze duplication already present
in pretraining data or evaluate a limited set of repetition settings.
Our study instead directly controls sensitive-data frequency during
fine-tuning across six levels, allowing us to measure how extraction
changes as repetition itself is varied.

\textbf{2.2 Privacy Leakage and Auditing During Fine-Tuning}

Privacy leakage can also be introduced during downstream fine-tuning.
Different adaptation methods exhibit different susceptibility to
membership inference and extraction attacks
\citep{mireshghallah2022empirical}. More recent work studies unintended PII leakage during fine-tuning and shows that factors such as PII frequency, task type, language, and model size can affect memorization
\citep{szep2026unintended}. Importantly, this work confirms that frequency matters even when sensitive information is introduced during adaptation, rather than only during large-scale pretraining.

Privacy leakage is measured using several auditing approaches. Membership-inference methods test whether an example was used in training,
and stronger recent attacks such as Window-Based Comparison exploit local
memorization signals to improve member detection
\citep{chen2026window}. Canary-based auditing likewise depends strongly on
which examples are chosen for the audit; stronger canary constructions can
reveal leakage that weaker canaries miss \citep{panda2025privacy}.
Per-document extraction results further show that aggregate membership
scores can hide leakage concentrated in individual records
\citep{maricato2026leakitperdocumentextraction}. Recent theoretical work also separates membership-style or memorization-based notions of privacy from direct extraction risk
\citep{liu2026beyond,che2026memorization}.

These results motivate evaluating both statistical memorization and practical extraction on the same sensitive examples. In our experiments, we measure statistical privacy signals and direct extraction, allowing us to test whether scores that detect memorization also identify which individual examples can actually be recovered.

\textbf{2.3 Privacy-Preserving Training}

Differentially private fine-tuning is a common approach for limiting the influence of individual training examples. Recent benchmarks evaluate
private LLM adaptation using both membership inference and canary
extraction, showing that practical privacy depends on the adaptation
setting and data distribution \citep{marek2026benchmarking}.

Gradient clipping can also reduce memorization without a formal privacy
guarantee. This behavior has been observed in language-model and
speech-recognition systems, and recent LoRA fine-tuning work combines clipping and Gaussian noise with other privacy mitigations \citep{huang2022detecting,wang2024unintended,bossy2025mitigating}.
These studies generally suggest that clipping can reduce memorization, but its behavior across different language models and different levels of sensitive-data repetition remains less clear.

\textbf{Positioning.}
Prior work establishes that repetition increases memorization
\citep{kandpal2022deduplicating,carlini2022quantifying},
that fine-tuning can introduce sensitive-data leakage
\citep{mireshghallah2022empirical,szep2026unintended},
that statistical privacy signals do not fully describe extraction risk
\citep{maricato2026leakitperdocumentextraction,
liu2026beyond,
che2026memorization},
and that privacy-preserving training can reduce leakage
\citep{marek2026benchmarking,huang2022detecting,
wang2024unintended,bossy2025mitigating}. Our study connects these questions in a single controlled experimental framework. We vary sensitive-data frequency from $1\times$ to $50\times$, measure memorization and direct extraction on the same examples, and compare matched non-private, clipping-only, and DP-SGD training across four language models. 
This design allows us to study not only whether repetition increases leakage, but also when sensitive information becomes recoverable and how privacy-preserving training changes that transition.

\section{Experimental Setup}
\label{sec:experimental-setup}

We design our experiments to study how the frequency of sensitive
information in the fine-tuning data affects memorization and extraction,
and how different privacy training methods change this behavior.
Unless otherwise stated, experiments use three independent training seeds,
$\{42,43,44\}$, and we report mean and standard deviation across seeds.
\subsection{Data and Tasks}

Our base fine-tuning data contain 75,000 examples from three tasks:
20,000 AG News examples for topic classification, 20,000 MNLI examples for natural-language inference, and 35,000 examples derived from AI4Privacy for PII extraction. AG News and MNLI provide general downstream tasks for measuring whether privacy training reduces model utility, while the PII task creates a setting in which the model learns to identify and generate sensitive values from text.

\textbf{Utility tasks.}
AG News is used for four-class topic classification, and MNLI is used for natural-language inference. These tasks do not contain the canaries used in our privacy evaluation and are used to measure downstream utility
before and after privacy-preserving training
\citep{zhang2015character,williams2018broad}.

\textbf{PII extraction task.}
We construct 35,000 training examples from AI4Privacy. Each example
contains a text passage with one or more pieces of personally identifiable
information and asks the model to return the PII values present in the
text. For example,
\begin{tcolorbox}[
    colback=blue!5,
    colframe=blue!50,
    arc=2pt,
    boxrule=0.5pt,
    left=10pt,right=10pt,top=4pt,bottom=4pt
]
\small
\textbf{Instruction:} Identify all personally identifiable information
in the following text. Return values separated by semicolons.\\
\textbf{Input:} Contact information for Jane Doe: 555-0199.\\
\textbf{Output:} \texttt{Jane Doe; 555-0199}
\end{tcolorbox}

This task is well suited to our study because the model is trained to generate PII as part of normal task completion. Our extraction experiments therefore test whether sensitive values seen during fine-tuning become recoverable in a task that already requires the model to produce such information, rather than introducing an unrelated extraction prompt.

\subsection{Canary Construction and Frequency Control}

We construct canaries from held-out AI4Privacy records. For each selected record, we identify an email address or phone number and split the text immediately before that value. The text before the split is used as the input context, while the email address or phone number is used as the sensitive target. For example,

\begin{tcolorbox}[
    colback=gray!8,
    colframe=black!40,
    arc=2pt,
    boxrule=0.5pt,
    left=10pt,right=10pt,top=4pt,bottom=4pt
]
\small
\textbf{Instruction:} Complete the missing private field from the text
below. Return only the value.\\
\textbf{Input:} Regarding the recent publication on radio wave
distribution patterns, please send all feedback to\\
\textbf{Output:} \texttt{NMBN@yahoo.com}
\end{tcolorbox}

\textbf{Members and references.}
Member and reference canaries are created using the same procedure and follow the same input and output format. They are drawn from disjoint AI4Privacy records. Member canaries are added to the fine-tuning data, while reference canaries and their source records are excluded from
training. This allows us to compare examples that were seen during fine-tuning with similar examples that were not seen.

\textbf{Frequency control.}
We define the training frequency of a canary as the number of times it appears in the fine-tuning data. Member canaries are assigned to one of six frequency levels,
\[
f \in \{1,2,5,10,20,50\}.
\]
The six groups contain 200, 150, 150, 120, 100, and 80 unique canaries,
respectively. Repeating each canary according to its assigned frequency
produces 8,450 canary training examples from 800 unique sensitive values.
We additionally construct 1,000 reference canaries that are never included
in training.

Different canaries are used at each frequency level. For example, a $50\times$ canary is not the same sensitive value used in the $1\times$ group. The experiment therefore compares groups of sensitive examples that appear at different frequencies, rather than repeatedly retraining the same secret at increasing frequencies.

\subsection{Distributional checks.}
\label{app:distribution-checks}
We also check whether member and reference canaries differ in simple properties that could affect memorization. We find no significant difference in target length using a two-sample Kolmogorov--Smirnov test ($D=0.042$, $p=0.416$), and no significant difference in email/phone composition using a $\chi^2$ test ($\chi^2=2.57$, $p=0.109$). These checks provide no evidence that these basic target properties explain the differences observed in our privacy results.
\subsection{Checkpoint selection}
\label{app:checkpoint-selection}
Training configurations and checkpoints are selected using validation utility rather than privacy outcomes. For each checkpoint, we compute
\[
U_{\mathrm{val}}
=
\frac{
A_{\mathrm{AG}}+
A_{\mathrm{MNLI}}+
F_{1,\mathrm{PII}}
}{3},
\]
where $A_{\mathrm{AG}}$ is AG News validation accuracy,
$A_{\mathrm{MNLI}}$ is MNLI validation accuracy, and
$F_{1,\mathrm{PII}}$ is validation F1 on the PII task. We select the
checkpoint with the highest $U_{\mathrm{val}}$.

Memorization, membership-inference, exposure, and extraction measurements are not used for hyperparameter or checkpoint selection. They are computed only after the final checkpoint has been fixed. This separation prevents privacy results from influencing model selection and allows the privacy--utility trade-off to be evaluated on independently selected models.
All models are fine-tuned using LoRA with rank $r=16$ and scaling parameter $\alpha=32$. LoRA is applied to the attention and feed-forward projection layers. We use each model's native tokenizer and chat template so that training and evaluation follow the format expected by the original model. For the controlled experiments, models are trained for three epochs with a learning rate of $10^{-5}$, AdamW optimization, bfloat16 precision, and a logical batch size of 64. Physical batch sizes and maximum sequence lengths are adjusted when needed to fit the memory requirements of each model, while the logical batch size is kept fixed.

Before the final privacy experiments, we evaluated multiple training configurations using downstream validation performance and training stability. The final configuration for each model was fixed before privacy evaluation. Memorization and extraction results were not used to select training hyperparameters.
\subsection{Models and Fine-Tuning}

We evaluate four language models: LLaMA-3.2-1B, Gemma-2-2B, Gemma-3-1B, and VaultGemma-1B. Using models from different families allows us to assess whether the level of repetition required for extraction is consistent across models or specific to a particular architecture.
We include VaultGemma-1B because it was pretrained from scratch with differential privacy. Built on the Gemma family, it provides a formal sequence-level privacy guarantee for its pretraining data \citep{sinha2025vaultgemma}. Including VaultGemma allows us to examine whether privacy-preserving pretraining affects how a model later responds to sensitive data introduced during downstream fine-tuning.

\subsection{Training Conditions}

We compare three training conditions to understand how privacy-preserving training changes memorization and extraction. The conditions are matched within each model so that the privacy mechanism is the main difference between them.


\textbf{Non-private training.} The non-private condition uses the same training pipeline as the privacy conditions but without per-example gradient clipping or added noise. This serves as the baseline for measuring how much information is memorized and extracted without a privacy mechanism.

\textbf{Clipping-only training.}
In the clipping-only condition, each example's gradient is clipped to a
maximum norm of $C=1.0$, but no noise is added. We include this condition because gradient clipping is one of the two main components of DP-SGD and can independently alter the optimization dynamics, potentially affecting memorization \citep{zhang2026understanding,ravikumar2026memorization}. Comparing non-private training with clipping-only training allows us to isolate the effect of clipping from the additional effect of DP noise.


\textbf{DP-SGD training.}
The DP-SGD condition uses the same clipping norm, $C=1.0$, and adds
Gaussian noise calibrated with Opacus to satisfy a target privacy budget\citep{abadi2016deep}.
For the controlled frequency experiments, we use
$\varepsilon=5$ and $\delta=10^{-5}$. Comparing clipping-only training
with DP-SGD lets us measure the additional change produced by adding DP
noise.

In addition to this controlled $\varepsilon=5$ experiment, we conduct a
broader privacy-budget audit at $\varepsilon\in\{3,5,8\}$ using the
$1\times$, $10\times$, and $50\times$ canary frequency levels. This
additional audit tests whether the observed privacy behavior is specific
to a single privacy budget.

All three matched conditions use the same three-epoch training horizon,
learning rate of $10^{-5}$, bfloat16 precision, 100 warmup steps, AdamW
optimization, and logical batch size of 64 within each model. Physical
microbatch size and sequence length are adjusted when needed for model
memory requirements, but are kept fixed across the three training
conditions for a given model.
\subsection{Privacy and Extraction Evaluation}
\label{sec:privacy-extraction}

We evaluate privacy leakage using both statistical privacy signals and generation-based extraction. The statistical metrics measure
whether training members behave differently from unseen references, whereas extraction measures whether the sensitive value can be
recovered from the model.

\textbf{Statistical privacy metrics.}
We first measure memorization using the member--reference loss gap,
defined as
\[
\mathrm{Gap}
=
\mathbb{E}_{z\in\mathcal{R}}[\ell(z)]
-
\mathbb{E}_{z\in\mathcal{M}}[\ell(z)],
\]
where $\mathcal{M}$ and $\mathcal{R}$ denote member and reference
canaries, respectively, and $\ell(z)$ is the language-model loss. A positive gap indicates that the model assigns lower loss to training members than to unseen references. Larger positive values indicate stronger memorization of the training members.

We additionally report the rank-based \emph{Exposure} score \citep{carlini2019secret} and Window-Based Comparison (WBC) membership-inference AUC \citep{chen2026window}. Higher Exposure indicates
that a canary is unusually well predicted relative to reference examples. For WBC, an AUC of $0.5$ corresponds to chance-level member--reference
distinguishability, while larger values indicate stronger membership signals. These metrics measure statistical evidence of memorization or training membership; they do not establish that the sensitive value can be recovered through generation.

\textbf{Threat model and extraction criterion.} We consider a black-box attacker with generation access to the fine-tuned model. The attacker knows the context preceding a sensitive value but not
the value itself, and has no access to model parameters, gradients, optimizer states, or the fine-tuning dataset. An extraction is counted as
successful when the target email address or phone number appears in the generated output. The target does not need to be the entire generated response, so extraction is intentionally distinct from utility exact match.

\textbf{Frequency-conditioned extraction.} For each training frequency $f$, we define the extraction rate as
\[
E(f)
=
\frac{
\text{number of extracted member canaries at frequency } f
}{
\text{number of evaluated member canaries at frequency } f
}.
\]

The values of $E(f)$ across
$f\in\{1,2,5,10,20,50\}$ form the extraction-response curve, which is our
primary measure of how practical leakage changes with repetition.

For compact comparison, we additionally define
\[
F^*=\min\{f:E(f)>0\}
\]
as the first tested frequency with any observed extraction, and
\[
F^*_{50}=\min\{f:E(f)\geq0.5\}
\]
as the first tested frequency at which at least half of the canaries are
extracted. Because $F^*$ can be determined by a single vulnerable
example, we use these thresholds only as descriptive summaries and treat
the full extraction-response curve as the primary result.

\subsection{Utility Evaluation}
\label{sec:utility}

Privacy protection is useful only if the fine-tuned model retains sufficient task performance. We therefore evaluate utility on the three tasks used during fine-tuning. For AG News and MNLI, we report
classification accuracy. For the PII task, we report both token-level F1 and exact match. Exact match requires the generated answer to match the
target output exactly, whereas F1 gives partial credit when only part of the expected PII is recovered. We additionally report language-model
perplexity as a general measure of predictive quality.

\section{Experimental Results}
\label{sec:results}
We evaluate privacy leakage from four perspectives:
frequency-conditioned extraction, matched clipping and DP-SGD comparisons, the relationship between statistical privacy metrics and practical extraction, and a broader audit across multiple privacy budgets and extraction attacks. 

\subsection{When Does Sensitive Information Become Extractable?}
\label{sec:dose-response}

\begin{figure}[t]
    \centering
    \includegraphics[width=\textwidth]
    {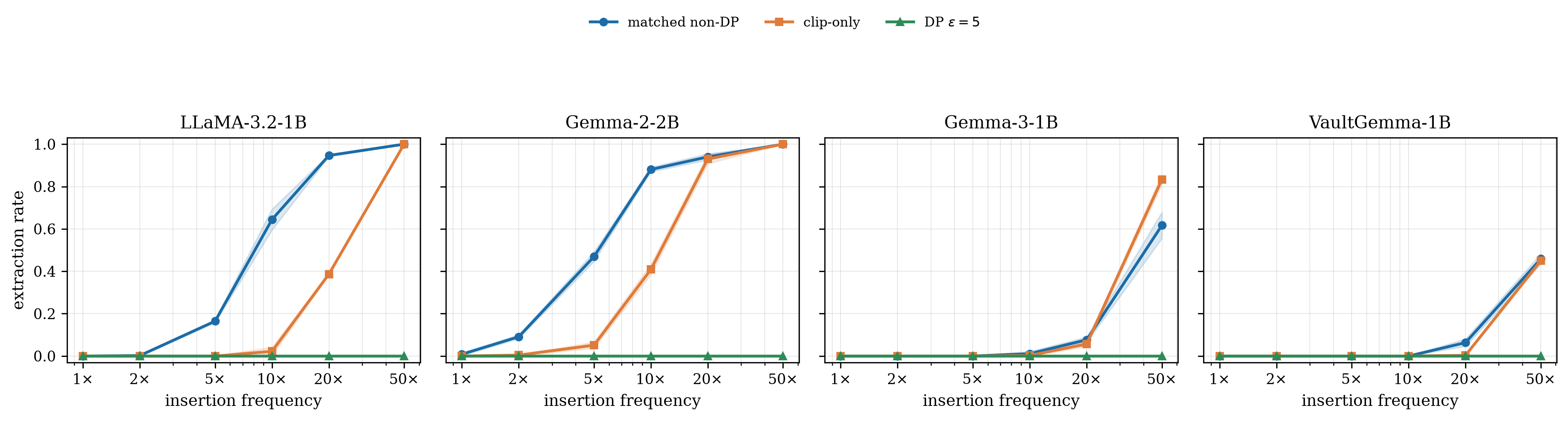}
    \caption{Frequency-conditioned all-canary greedy extraction across four
    language models under matched non-private, clipping-only, and DP-SGD
    training. Points show means across three seeds; shaded regions indicate
    $\pm 1$ standard deviation.}
    \label{fig:extraction_curves}
\end{figure}

We first examine how extraction changes as sensitive information appears more frequently in the fine-tuning data. Figure~\ref{fig:extraction_curves}
shows all-canary greedy extraction across the six controlled frequencies $f\in\{1,2,5,10,20,50\}$. We begin with the matched non-private condition
to isolate the relationship between repetition and extraction before examining privacy-preserving training.

\textbf{Extraction increases with frequency, but the transition differs substantially across models.}
As shown in Figure~\ref{fig:extraction_curves}, with full numerical results
reported in Appendix
Table~\ref{tab:full-extraction}, LLaMA-3.2-1B shows little extraction at low frequencies, but the rate rises from $16.4\%$ at $5\times$ to $64.4\%$ at $10\times$ and $94.7\%$ at
$20\times$. Gemma-2-2B transitions even earlier, reaching $46.9\%$ extraction at $5\times$ and $88.1\%$ at $10\times$. The transition occurs later for Gemma-3-1B and VaultGemma-1B. Gemma-3 remains near zero through $10\times$, then increases to $7.7\%$ at $20\times$ and $61.7\%$ at $50\times$. VaultGemma shows the latest transition in our tested range, with $6.3\%$ extraction at $20\times$
and $45.8\%$ at $50\times$.


These results show that repetition strongly affects whether sensitive information becomes recoverable, but the frequency at which extraction begins varies across models. Consequently, the same sensitive-data frequency can correspond to substantially different levels of practical leakage across models.

\textbf{Threshold summaries highlight differences across models and runs.}
We summarize the extraction curves using $F^*_{50}$, the first tested frequency at which at least half of the member canaries are extracted. Table~\ref{tab:f50-main} reports this threshold for the three training conditions. Full $F^*$ and $F^*_{50}$ ranges across seeds are reported in Appendix Table ~\ref{tab:full-thresholds}.
\begin{table}[t]
\centering
\caption{First tested frequency at which at least 50\% of member
canaries are extracted ($F^*_{50}$). N.O. indicates that the threshold
was not observed through $50\times$.}
\label{tab:f50-main}
\small
\begin{tabular}{lccc}
\toprule
Model & Non-DP & Clipping & DP-SGD \\
\midrule
LLaMA-3.2-1B  & $10\times$ & \textbf{$50\times$} & N.O. \\
Gemma-2-2B    & $10\times$ & \textbf{$20\times$} & N.O. \\
Gemma-3-1B    & $50\times$ & $50\times$ & N.O. \\
VaultGemma-1B & N.O.       & N.O.       & N.O. \\
\bottomrule
\end{tabular}
\end{table}
Under matched non-private training, LLaMA and Gemma-2 reach majority extraction by $10\times$, whereas Gemma-3 reaches this level only at $50\times$. VaultGemma does not reach $F^*_{50}$ within the tested range.
The full appendix
Table~\ref{tab:full-thresholds} results also show that $F^*$ varies across seeds for several models, indicating that the first observed extraction event can depend on the particular training run. For this reason, we treat the full
extraction-response curve as the primary result and use $F^*$ and
$F^*_{50}$ only as descriptive summaries.

\textbf{Within-Canary Validation of the Frequency Effect.}
To verify that the observed frequency trend is not driven only by differences between canaries assigned to different frequency groups, we ran an additional within-canary experiment on Gemma-2-2B using the same 50 canaries at $1\times$, $5\times$, $10\times$, $20\times$, and $50\times$ repetition across three seeds. Mean greedy extraction increased from $0.00$ at $1\times$ to $0.36$ at $5\times$, $0.90$ at $10\times$, $0.95$ at $20\times$, and $0.98$ at $50\times$. Overall, 49 of 50 canaries changed from non-extracted at low frequency to extracted by $50\times$, and 98\% of canary--seed trajectories were non-decreasing with frequency. This within-canary result confirms that increasing repetition itself strongly increases extraction probability in this setting.

\subsection{How Do Gradient Clipping and DP Noise Change Extraction?}
\label{sec:mechanisms}

We next examine how the extraction curves change under clipping-only and
DP-SGD training. Because the three conditions are matched within each
model, this comparison separates the change associated with gradient
clipping from the additional change produced by DP noise. 
\begin{table*}[!t]
\centering
\caption{Matched comparison of non-private, clipping-only, and DP-SGD
training. Results are mean $\pm$ standard deviation over three seeds.
Gap measures member--reference loss separation, and WBC AUC measures
membership inference.}
\label{tab:mechanism-comparison}
\small
\begin{tabular}{llccccc}
\toprule
Model & Training &
AG $\uparrow$ &
MNLI $\uparrow$ &
PII F1 $\uparrow$ &
Gap &
WBC AUC \\
\midrule

LLaMA-3.2-1B
& matched non-DP
& .899$\pm$.001
& .788$\pm$.005
& .989$\pm$.001
& .358$\pm$.000
& .631$\pm$.038 \\
& clip-only
& .906$\pm$.001
& .807$\pm$.004
& .990$\pm$.000
& .211$\pm$.009
& .563$\pm$.004 \\
& DP $\varepsilon=5$
& .778$\pm$.004
& .474$\pm$.007
& .789$\pm$.002
& \textbf{.018$\pm$.001}
& \textbf{.453$\pm$.002} \\
\midrule

Gemma-2-2B
& matched non-DP
& .901$\pm$.001
& .865$\pm$.001
& .993$\pm$.000
& .638$\pm$.004
& .738$\pm$.011 \\
& clip-only
& .906$\pm$.001
& .863$\pm$.002
& .995$\pm$.000
& .594$\pm$.029
& .706$\pm$.029 \\
& DP $\varepsilon=5$
& .833$\pm$.001
& .738$\pm$.003
& .828$\pm$.001
& \textbf{-.006$\pm$.000}
& \textbf{.497$\pm$.003} \\
\midrule

Gemma-3-1B
& matched non-DP
& .896$\pm$.004
& .746$\pm$.004
& .967$\pm$.002
& .227$\pm$.023
& .514$\pm$.006 \\
& clip-only
& .897$\pm$.001
& .783$\pm$.002
& .975$\pm$.006
& {.250$\pm$.015}
& {.528$\pm$.008} \\
& DP $\varepsilon=5$
& .747$\pm$.001
& .580$\pm$.002
& .745$\pm$.002
& \textbf{-.012$\pm$.001}
& \textbf{.492$\pm$.006} \\
\midrule

VaultGemma-1B
& matched non-DP
& .887$\pm$.001
& .541$\pm$.004
& .965$\pm$.001
& .212$\pm$.006
& .697$\pm$.002 \\
& clip-only
& .891$\pm$.002
& .639$\pm$.003
& .973$\pm$.001
& .178$\pm$.002
& .661$\pm$.002 \\
& DP $\varepsilon=5$
& .478$\pm$.006
& .324$\pm$.007
& .634$\pm$.013
& \textbf{-.026$\pm$.001}
& \textbf{.470$\pm$.003} \\
\bottomrule
\end{tabular}
\vspace{-1.5em}
\end{table*}

\textbf{Gradient clipping delays extraction for some models, but the effect is not uniform.} As shown in Figure~\ref{fig:extraction_curves}, with full numerical results reported in Appendix Table~\ref{tab:full-extraction}, clipping shifts the extraction
curve substantially to the right for LLaMA-3.2-1B.  Extraction remains near zero through $10\times$, rises to $38.7\%$ at $20\times$, and reaches $100\%$ at $50\times$. This moves $F^*_{50}$ from $10\times$ under matched non-private training to $50\times$ under clipping. Gemma-2-2B shows a similar but smaller shift. Under clipping, extraction is only $5.1\%$ at $5\times$ and $40.8\%$ at $10\times$, compared with
$46.9\%$ and $88.1\%$, respectively, under matched non-private training. Its $F^*_{50}$ therefore moves from $10\times$ to $20\times$. (Table~\ref{tab:f50-main})

The effect is weaker for VaultGemma-1B and differs for Gemma-3-1B.
VaultGemma remains nearly unchanged at high frequency, reaching
approximately $45\%$ extraction at $50\times$ under both matched
non-private and clipping-only training. Gemma-3 provides a different
pattern: clipping produces $83.3\%$ extraction at $50\times$, compared
with $61.7\%$ under matched non-private training. Thus, clipping can delay
extraction substantially for some models, but it does not provide a consistent reduction across models or frequencies.


\textbf{DP-SGD produces a stronger and more consistent reduction.} Under DP-SGD at $\varepsilon=5$, we observe no successful all-canary
greedy extractions at any tested frequency for any of the four models. This corresponds to $0/9{,}600$ successful extractions across models, seeds, and frequency levels. The same result holds under the repeated stochastic evaluation: no member canary is recovered in any of the $K=10$ sampling trials (Appendix Table~\ref{tab:full-extraction-stochastic}).

These results indicate that adding calibrated DP noise changes the extraction behavior more strongly and consistently than clipping alone within the tested setting. Importantly, zero observed extraction should not be interpreted as evidence that extraction is impossible; rather, this result is specific to the attacks and frequency range evaluated here.

\textbf{Changes in extraction are accompanied by different utility costs.} Table~\ref{tab:mechanism-comparison} compares downstream utility and statistical privacy metrics for the three training conditions. Clipping generally preserves task performance while reducing leakage for some models. DP-SGD produces much larger reductions in memorization and membership-inference signals, but also introduces larger and model-dependent utility losses. For example, MNLI accuracy decreases from
$0.807$ under clipping to $0.474$ under DP for LLaMA-3.2-1B, from $0.863$ to $0.738$ for Gemma-2-2B, and from $0.639$ to $0.324$ for VaultGemma-1B.

The comparison therefore shows two distinct behaviors: clipping can shift the frequency at which extraction becomes likely without consistently
preventing high-frequency leakage, whereas DP-SGD provides a much stronger reduction in observed extraction at the cost of greater utility degradation.

\subsection{Do Statistical Privacy Metrics Predict Extraction?}
\label{sec:metric-extraction}

We next ask whether statistical privacy scores can identify which individual sensitive examples become practically extractable. This is different from conventional membership inference, which asks whether an example was present in the training data. Here, we restrict the analysis to member canaries and ask whether their statistical privacy scores distinguish examples that are successfully extracted from those that are not.

We evaluate three per-canary scores: loss-based memorization, WBC, and the rank-based Exposure score. For each score, we compute AUROC using
all-canary greedy extraction as the binary outcome (Table~\ref{tab:metric-extraction}). An AUROC of $0.5$ indicates chance-level separation between extracted and non-extracted
members, while $1.0$ indicates perfect separation. DP-SGD is excluded because no successful extraction events are observed (Appendix Table~\ref{tab:full-extraction}), making extraction-prediction AUROC undefined.

\begin{table}[!t]
\centering
\caption{Ability of statistical privacy scores to predict extraction among member canaries. AUROC uses all-canary greedy extraction as the binary outcome and is pooled across three training seeds. An AUROC of $0.5$ corresponds to chance-level separation. DP-SGD is omitted because no successful extraction events are observed, making extraction-prediction
AUROC undefined.}
\label{tab:metric-extraction}
\small
\begin{tabular}{llcccc}
\toprule
Model & Training &
Extracted / Total &
Loss AUROC &
WBC AUROC &
Exposure AUROC \\
\midrule

LLaMA-3.2-1B
& matched non-DP
& 831 / 2400
& .695
& .519
& .695 \\
& clip-only
& 364 / 2400
& .748
& .571
& .749 \\
\midrule

Gemma-2-2B
& matched non-DP
& 1096 / 2400
& .578
& \textbf{.468}
& .578 \\
& clip-only
& 691 / 2400
& .693
& .584
& .694 \\
\midrule

Gemma-3-1B
& matched non-DP
& 175 / 2400
& .641
& .508
& .641 \\
& clip-only
& 218 / 2400
& .734
& .577
& .734 \\
\midrule

VaultGemma-1B
& matched non-DP
& 129 / 2400
& .508
& .661
& .508 \\
& clip-only
& 109 / 2400
& .666
& \textbf{.791}
& .666 \\
\midrule
\textbf{Pooled}
& \textbf{{non-DP + clip-only}}
& \textbf{3613 / 19200}
& \textbf{.691}
& \textbf{.544}
& \textbf{.689} \\
\bottomrule
\end{tabular}
\vspace{-1.5em}
\end{table}

\textbf{Strong membership signals do not necessarily predict extraction.}
Across models and non-DP training conditions, statistical privacy scores
are substantially less reliable at predicting extraction than they are at
distinguishing training members from references. Pooled across the matched
non-private and clipping-only conditions, loss-based memorization achieves
an extraction-prediction AUROC of $0.691$, the Exposure score achieves
$0.689$, and WBC reaches only $0.544$ (Table~\ref{tab:metric-extraction}).

This is particularly notable for WBC. In the broader membership-inference audit, which uses different model-specific training configurations, several non-private models achieve WBC AUCs above $0.96$ when distinguishing members from references
(Appendix Table~\ref{tab:broad-privacy}). However, when the task changes to predicting which members will actually be extracted, the pooled WBC AUROC falls close to chance. A strong signal that an example influenced the model therefore does not necessarily identify whether its sensitive value will be reproduced during generation.

\textbf{The relationship between privacy scores and extraction is model dependent.}
Prediction performance also varies substantially across models and training conditions. Under matched non-private training, WBC achieves an AUROC of only $0.468$ for Gemma-2-2B, while under clipping it reaches $0.791$ for VaultGemma-1B. Loss-based memorization and Exposure are more consistent overall, but neither provides uniformly strong prediction across all settings  (Table~\ref{tab:metric-extraction}).

No evaluated statistical privacy metric is uniformly predictive of per-canary extraction across the models and training conditions in our study. These results support reporting direct extraction alongside
statistical privacy metrics. Statistical memorization and membership signals remain useful for detecting whether training data have influenced a model, but direct
extraction evaluation is still necessary when the goal is to measure whether sensitive information can actually be recovered.

\subsection{Robustness Across Privacy Budgets and Extraction Attacks}
\label{sec:robustness}

We additionally test whether the observed privacy behavior is specific to
$\varepsilon=5$ or to greedy decoding. Across
$\varepsilon\in\{3,5,8\}$, WBC remains near chance and we observe no
successful extraction at $1\times$, $10\times$, or $50\times$ for any
model. Repeating extraction with $K=10$ stochastic generations per canary
at $\varepsilon=5$ likewise yields zero observed extraction across all
models and tested frequencies. 

\textbf{Privacy leakage remains low across the tested DP budgets.}\label{app:broad-audit} Across $\varepsilon=3$, $5$, and $8$, WBC membership-inference AUC remains close to chance for all four models, generally around $0.49$--$0.50$. The member--reference memorization gap is also substantially smaller than
under non-private fine-tuning. We observe no successful extraction at $1\times$, $10\times$, or $50\times$ for any model under any of the
tested DP budgets (Appendix Table~\ref{tab:broad-privacy}).

We therefore observe little empirical separation among $\varepsilon=3$, $5$, and $8$ on the privacy measurements considered here. Within the models, training horizon, frequencies, and attacks evaluated in our experiments, the transition from non-private to DP training produces a much larger observable change than the differences among these three privacy budgets.
\subsection{Stochastic Extraction Robustness}
\label{app:stochastic-extraction}

To test whether the extraction results depend on deterministic greedy decoding, we repeat the all-canary extraction evaluation using stochastic
sampling with $K=10$ generations per canary. We use temperature $=1.0$,
top-$p=1.0$, and report whether each canary is recovered in at least one
of the ten generations. Table~\ref{tab:full-extraction-stochastic} reports
the complete results.

The stochastic results broadly follow the same frequency-dependent patterns
as the greedy evaluation in Table~\ref{tab:full-extraction}. Extraction
increases with repetition for the non-private and clipping-only conditions,
while DP-SGD at $\varepsilon=5$ produces zero observed extraction at every
tested frequency across all four models. At very low extraction rates,
occasional stochastic hits can produce small differences from greedy
decoding. For example, Gemma-3-1B shows rates of $0.002$ in a few
$1\times$ or $2\times$ cells under stochastic sampling even though the
corresponding greedy rates are zero. These values correspond to a small
number of isolated sampled recoveries rather than a systematic change in
the extraction pattern.


\section{Discussion}

Our results show that practical extraction risk depends strongly on
sensitive-data frequency and model choice. The same repetition level can
produce near-zero extraction for one model and substantial extraction for
another, making the full extraction-response curve more informative than
a single aggregate memorization score or threshold.

The matched experiments also show that clipping and DP noise should not
be treated as interchangeable. Clipping can delay extraction, but its
effect is model dependent and does not consistently suppress
high-frequency leakage. DP-SGD produces a much stronger reduction under
the tested conditions, while also incurring larger utility costs.
Statistical membership and memorization signals likewise do not reliably
identify which individual examples become extractable, supporting direct
extraction evaluation as a complement to aggregate privacy metrics.
\section{Limitations}

Our study uses controlled email- and phone-number canaries derived from PII-oriented data, so the observed extraction curves may not generalize
directly to other types of sensitive information. Exposure groups contain different canaries rather than repeatedly increasing the frequency of the
same secret across a sequence of models, so the dose-response analysis is population-level rather than longitudinal. We evaluate LoRA adaptation of
approximately 1--2B parameter models; larger models, full-model fine-tuning, and proprietary systems may exhibit different behavior. Our extraction analysis is also limited to the tested greedy and stochastic
attacks, and stronger adaptive prompting or larger sampling budgets could reveal additional leakage. Finally, the broad audit uses model-specific
utility-selected training configurations, so controlled mechanism conclusions rely primarily on the matched dose-response experiments.

\section{Conclusion}

We study when practical extraction begins to emerge as sensitive
information appears more frequently during downstream fine-tuning, and
how this behavior relates to statistical memorization. Extraction increases sharply with repetition, but the location of this transition varies substantially across models and training runs. Matched comparisons show that gradient clipping can delay extraction but
does not provide uniform protection at high exposure, whereas DP-SGD at $\varepsilon=5$ yields zero observed extraction through $50\times$ under
the tested greedy and stochastic attacks. We also find that statistical memorization and membership-inference scores are not uniformly reliable
predictors of which individual examples become extractable. These results support exposure-conditioned direct extraction auditing as a useful
complement to aggregate privacy metrics.

\bibliographystyle{unsrtnat}
\bibliography{references} 

\newpage
\appendix

\section{Technical appendices and supplementary material} \label{additional}

\begin{table}[!htp]
\centering
\caption{Downstream utility across models and privacy budgets. Results are mean $\pm$ standard deviation over three seeds.}
\label{tab:broad-utility}
\small
\resizebox{\textwidth}{!}{
\begin{tabular}{llccccc}
\toprule
Model & Training & AG $\uparrow$ & MNLI $\uparrow$ &
PII F1 $\uparrow$ & PII EM $\uparrow$ & PPL $\downarrow$ \\
\midrule

LLaMA-3.2-1B
& non-DP & 0.901 $\pm$ 0.003 & 0.795 $\pm$ 0.007 & 0.987 $\pm$ 0.001 & 0.946 $\pm$ 0.002 & 1.065 $\pm$ 0.001 \\
& $\epsilon=3$ & 0.753 $\pm$ 0.003 & 0.462 $\pm$ 0.009 & 0.770 $\pm$ 0.000 & 0.442 $\pm$ 0.002 & 1.394 $\pm$ 0.001 \\
& $\epsilon=5$ & 0.760 $\pm$ 0.004 & 0.472 $\pm$ 0.001 & 0.783 $\pm$ 0.002 & 0.459 $\pm$ 0.004 & 1.381 $\pm$ 0.000 \\
& $\epsilon=8$ & 0.771 $\pm$ 0.004 & 0.483 $\pm$ 0.001 & 0.792 $\pm$ 0.001 & 0.473 $\pm$ 0.001 & 1.373 $\pm$ 0.001 \\
\midrule

Gemma-2-2B
& non-DP & 0.926 $\pm$ 0.002 & 0.872 $\pm$ 0.002 & 0.995 $\pm$ 0.001 & 0.977 $\pm$ 0.003 & 1.043 $\pm$ 0.002 \\
& $\epsilon=3$ & 0.871 $\pm$ 0.002 & 0.789 $\pm$ 0.001 & 0.958 $\pm$ 0.001 & 0.848 $\pm$ 0.002 & 1.131 $\pm$ 0.002 \\
& $\epsilon=5$ & 0.872 $\pm$ 0.002 & 0.792 $\pm$ 0.002 & 0.962 $\pm$ 0.001 & 0.863 $\pm$ 0.001 & 1.129 $\pm$ 0.002 \\
& $\epsilon=8$ & 0.874 $\pm$ 0.002 & 0.796 $\pm$ 0.003 & 0.964 $\pm$ 0.001 & 0.869 $\pm$ 0.001 & 1.127 $\pm$ 0.001 \\
\midrule

Gemma-3-1B
& non-DP & 0.916 $\pm$ 0.005 & 0.806 $\pm$ 0.003 & 0.986 $\pm$ 0.001 & 0.950 $\pm$ 0.002 & 1.053 $\pm$ 0.000 \\
& $\epsilon=3$ & 0.858 $\pm$ 0.001 & 0.720 $\pm$ 0.002 & 0.901 $\pm$ 0.000 & 0.680 $\pm$ 0.001 & 1.195 $\pm$ 0.002 \\
& $\epsilon=5$ & 0.860 $\pm$ 0.001 & 0.724 $\pm$ 0.001 & 0.909 $\pm$ 0.004 & 0.697 $\pm$ 0.011 & 1.190 $\pm$ 0.001 \\
& $\epsilon=8$ & 0.862 $\pm$ 0.001 & 0.729 $\pm$ 0.002 & 0.920 $\pm$ 0.000 & 0.727 $\pm$ 0.000 & 1.180 $\pm$ 0.003 \\
\midrule

VaultGemma-1B
& non-DP & 0.907 $\pm$ 0.002 & 0.599 $\pm$ 0.000 & 0.974 $\pm$ 0.001 & 0.922 $\pm$ 0.004 & 1.099 $\pm$ 0.001 \\
& $\epsilon=3$ & 0.825 $\pm$ 0.004 & 0.363 $\pm$ 0.011 & 0.852 $\pm$ 0.005 & 0.595 $\pm$ 0.010 & 1.275 $\pm$ 0.002 \\
& $\epsilon=5$ & 0.833 $\pm$ 0.009 & 0.367 $\pm$ 0.011 & 0.859 $\pm$ 0.004 & 0.613 $\pm$ 0.007 & 1.270 $\pm$ 0.002 \\
& $\epsilon=8$ & 0.839 $\pm$ 0.003 & 0.377 $\pm$ 0.017 & 0.866 $\pm$ 0.004 & 0.628 $\pm$ 0.010 & 1.265 $\pm$ 0.004 \\
\bottomrule
\end{tabular}}
\end{table}

\begin{figure}[!ht]
    \centering
    \includegraphics[width=\textwidth]
    {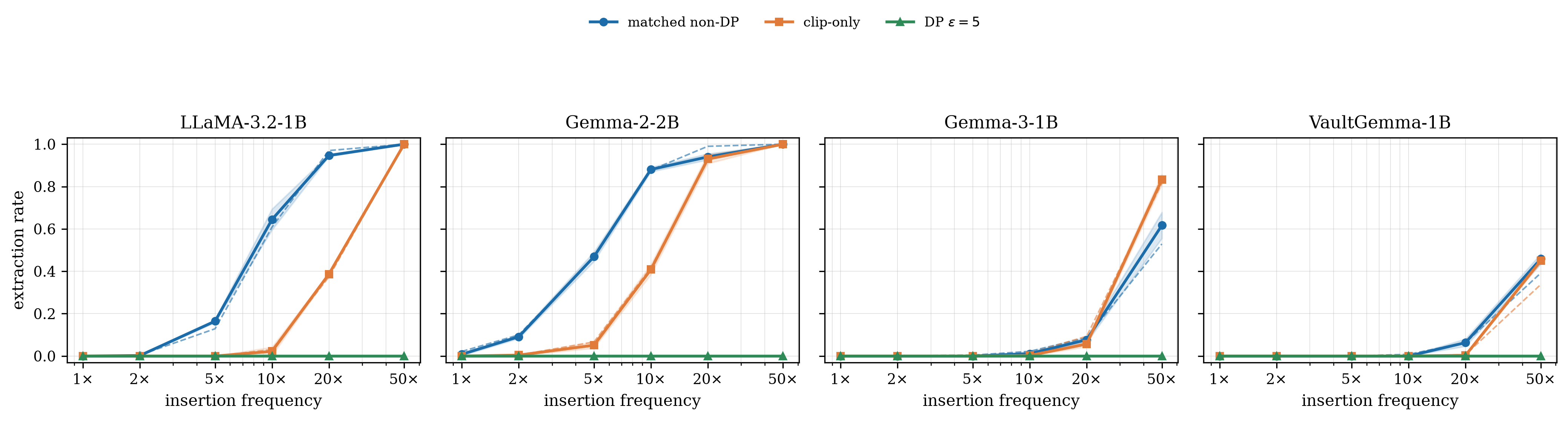}
    \caption{
    Comparison of all-canary greedy and stochastic $K=10$ extraction
    across exposure levels. The stochastic extraction curves closely
    follow the greedy results, indicating that the primary
    frequency-conditioned trends are not specific to deterministic
    decoding.
    }
    \label{fig:extraction-curves-stochastic}
\end{figure}

\begin{table}[htp]
\centering
\caption{All-canary greedy extraction rate by insertion frequency, for
each model and training condition. Results are mean $\pm$ standard
deviation over three seeds. This is the full data underlying
Figure~\ref{fig:extraction_curves}.}
\label{tab:full-extraction}
\small
\resizebox{\linewidth}{!}{
\begin{tabular}{llcccccc}
\toprule
Model & Training & $1\times$ & $2\times$ & $5\times$ & $10\times$ & $20\times$ & $50\times$ \\
\midrule

LLaMA-3.2-1B
& matched non-DP & 0.000$\pm$0.000 & 0.002$\pm$0.003 & 0.164$\pm$0.006 & 0.644$\pm$0.049 & 0.947$\pm$0.005 & 1.000$\pm$0.000 \\
& clip-only       & 0.000$\pm$0.000 & 0.000$\pm$0.000 & 0.000$\pm$0.000 & 0.022$\pm$0.020 & 0.387$\pm$0.005 & 1.000$\pm$0.000 \\
& DP $\varepsilon=5$  & 0.000$\pm$0.000 & 0.000$\pm$0.000 & 0.000$\pm$0.000 & 0.000$\pm$0.000 & 0.000$\pm$0.000 & 0.000$\pm$0.000 \\
\midrule

Gemma-2-2B
& matched non-DP & 0.008$\pm$0.006 & 0.091$\pm$0.008 & 0.469$\pm$0.022 & 0.881$\pm$0.010 & 0.940$\pm$0.014 & 1.000$\pm$0.000 \\
& clip-only       & 0.000$\pm$0.000 & 0.004$\pm$0.003 & 0.051$\pm$0.014 & 0.408$\pm$0.024 & 0.930$\pm$0.022 & 1.000$\pm$0.000 \\
& DP $\varepsilon=5$  & 0.000$\pm$0.000 & 0.000$\pm$0.000 & 0.000$\pm$0.000 & 0.000$\pm$0.000 & 0.000$\pm$0.000 & 0.000$\pm$0.000 \\
\midrule

Gemma-3-1B
& matched non-DP & 0.000$\pm$0.000 & 0.000$\pm$0.000 & 0.000$\pm$0.000 & 0.011$\pm$0.010 & 0.077$\pm$0.012 & 0.617$\pm$0.062 \\
& clip-only       & 0.000$\pm$0.000 & 0.000$\pm$0.000 & 0.000$\pm$0.000 & 0.003$\pm$0.004 & 0.057$\pm$0.012 & 0.833$\pm$0.024 \\
& DP $\varepsilon=5$  & 0.000$\pm$0.000 & 0.000$\pm$0.000 & 0.000$\pm$0.000 & 0.000$\pm$0.000 & 0.000$\pm$0.000 & 0.000$\pm$0.000 \\
\midrule

VaultGemma-1B
& matched non-DP & 0.000$\pm$0.000 & 0.000$\pm$0.000 & 0.000$\pm$0.000 & 0.000$\pm$0.000 & 0.063$\pm$0.017 & 0.458$\pm$0.024 \\
& clip-only       & 0.000$\pm$0.000 & 0.000$\pm$0.000 & 0.000$\pm$0.000 & 0.000$\pm$0.000 & 0.003$\pm$0.005 & 0.450$\pm$0.010 \\
& DP $\varepsilon=5$  & 0.000$\pm$0.000 & 0.000$\pm$0.000 & 0.000$\pm$0.000 & 0.000$\pm$0.000 & 0.000$\pm$0.000 & 0.000$\pm$0.000 \\
\bottomrule
\end{tabular}}
\end{table}

\begin{table}[!htp]
\centering
\caption{Extraction-threshold summaries across three training seeds.
$F^*$ is the first tested frequency with any observed extraction, and
$F^*_{50}$ is the first tested frequency at which at least 50\% of member
canaries are extracted. Ranges indicate variation across seeds. N.O.
indicates that the threshold was not observed through $50\times$.}
\label{tab:full-thresholds}
\small
\begin{tabular}{llcc}
\toprule
Model & Training & $F^*$ & $F^*_{50}$ \\
\midrule

LLaMA-3.2-1B
& Matched non-DP & $2$--$5\times$ & $10\times$ \\
& Clipping-only  & $10\times$     & \textbf{$50\times$} \\
& DP-SGD         & N.O.            & N.O. \\

\midrule

Gemma-2-2B
& Matched non-DP & $1$--$2\times$ & $10\times$ \\
& Clipping-only  & $2$--$5\times$ & \textbf{$20\times$} \\
& DP-SGD         & N.O.            & N.O. \\

\midrule

Gemma-3-1B
& Matched non-DP & $10$--$20\times$ & $50\times$ \\
& Clipping-only  & $10$--$20\times$ & $50\times$ \\
& DP-SGD         & N.O.              & N.O. \\

\midrule

VaultGemma-1B
& Matched non-DP & $20\times$        & N.O. \\
& Clipping-only  & $20$--$50\times$ & N.O. \\
& DP-SGD         & N.O.               & N.O. \\

\bottomrule
\end{tabular}
\end{table}

\begin{table}[!ht]
\centering
\caption{Privacy metrics across models and privacy budgets. Extraction
rates are all-canary greedy extraction at each exposure tier. Results are
mean $\pm$ standard deviation over three seeds.}
\label{tab:broad-privacy}
\small
\resizebox{0.9\textwidth}{!}{
\begin{tabular}{llccccccc}
\toprule
Model & Training & Gap & WBC AUC & Exposure & Extr.\ 1$\times$ & Extr.\ 10$\times$ & Extr.\ 50$\times$ \\
\midrule

LLaMA-3.2-1B
& non-DP & 0.273 $\pm$ 0.004 & 0.660 $\pm$ 0.013 & 2.509 $\pm$ 0.022 & 0.003 $\pm$ 0.001 & 0.693 $\pm$ 0.020 & 0.944 $\pm$ 0.010 \\
& $\epsilon=3$ & 0.002 $\pm$ 0.000 & 0.493 $\pm$ 0.002 & 1.467 $\pm$ 0.001 & 0.000 $\pm$ 0.000 & 0.000 $\pm$ 0.000 & 0.000 $\pm$ 0.000 \\
& $\epsilon=5$ & 0.002 $\pm$ 0.000 & 0.495 $\pm$ 0.002 & 1.467 $\pm$ 0.001 & 0.000 $\pm$ 0.000 & 0.000 $\pm$ 0.000 & 0.000 $\pm$ 0.000 \\
& $\epsilon=8$ & 0.002 $\pm$ 0.000 & 0.491 $\pm$ 0.001 & 1.469 $\pm$ 0.001 & 0.000 $\pm$ 0.000 & 0.000 $\pm$ 0.000 & 0.000 $\pm$ 0.000 \\
\midrule

Gemma-2-2B
& non-DP & 2.481 $\pm$ 0.061 & 0.981 $\pm$ 0.001 & 8.154 $\pm$ 0.178 & 0.074 $\pm$ 0.025 & 0.944 $\pm$ 0.011 & 0.975 $\pm$ 0.000 \\
& $\epsilon=3$ & 0.069 $\pm$ 0.001 & 0.494 $\pm$ 0.003 & 1.650 $\pm$ 0.004 & 0.000 $\pm$ 0.000 & 0.000 $\pm$ 0.000 & 0.000 $\pm$ 0.000 \\
& $\epsilon=5$ & 0.069 $\pm$ 0.001 & 0.494 $\pm$ 0.002 & 1.631 $\pm$ 0.002 & 0.000 $\pm$ 0.000 & 0.000 $\pm$ 0.000 & 0.000 $\pm$ 0.000 \\
& $\epsilon=8$ & 0.070 $\pm$ 0.001 & 0.494 $\pm$ 0.002 & 1.623 $\pm$ 0.004 & 0.000 $\pm$ 0.000 & 0.000 $\pm$ 0.000 & 0.000 $\pm$ 0.000 \\
\midrule

Gemma-3-1B
& non-DP & 3.992 $\pm$ 0.777 & 0.965 $\pm$ 0.018 & 9.026 $\pm$ 0.981 & 0.298 $\pm$ 0.204 & 0.936 $\pm$ 0.035 & 0.972 $\pm$ 0.004 \\
& $\epsilon=3$ & 0.070 $\pm$ 0.001 & 0.489 $\pm$ 0.001 & 1.632 $\pm$ 0.003 & 0.000 $\pm$ 0.000 & 0.000 $\pm$ 0.000 & 0.000 $\pm$ 0.000 \\
& $\epsilon=5$ & 0.070 $\pm$ 0.002 & 0.489 $\pm$ 0.002 & 1.629 $\pm$ 0.005 & 0.000 $\pm$ 0.000 & 0.000 $\pm$ 0.000 & 0.000 $\pm$ 0.000 \\
& $\epsilon=8$ & 0.068 $\pm$ 0.003 & 0.489 $\pm$ 0.000 & 1.627 $\pm$ 0.012 & 0.000 $\pm$ 0.000 & 0.000 $\pm$ 0.000 & 0.000 $\pm$ 0.000 \\
\midrule

VaultGemma-1B
& non-DP & 3.718 $\pm$ 0.058 & 0.984 $\pm$ 0.002 & 8.967 $\pm$ 0.084 & 0.044 $\pm$ 0.016 & 0.958 $\pm$ 0.003 & 0.975 $\pm$ 0.000 \\
& $\epsilon=3$ & 0.066 $\pm$ 0.004 & 0.502 $\pm$ 0.006 & 1.612 $\pm$ 0.007 & 0.000 $\pm$ 0.000 & 0.000 $\pm$ 0.000 & 0.000 $\pm$ 0.000 \\
& $\epsilon=5$ & 0.070 $\pm$ 0.004 & 0.502 $\pm$ 0.005 & 1.627 $\pm$ 0.015 & 0.000 $\pm$ 0.000 & 0.000 $\pm$ 0.000 & 0.000 $\pm$ 0.000 \\
& $\epsilon=8$ & 0.075 $\pm$ 0.003 & 0.504 $\pm$ 0.004 & 1.642 $\pm$ 0.012 & 0.000 $\pm$ 0.000 & 0.000 $\pm$ 0.000 & 0.000 $\pm$ 0.000 \\
\bottomrule
\end{tabular}}
\end{table}

\subsection{Full Greedy Extraction Results}
\label{app:full-greedy}

Table~\ref{tab:full-extraction} reports the complete all-canary greedy extraction results underlying Figure~\ref{fig:extraction_curves}. The table
shows extraction rates at all six controlled insertion frequencies for each model and training condition. These values provide the exact numerical results used to construct the frequency-conditioned extraction curves in the main paper.

\begin{table}[!htp]
\centering
\caption{Stochastic $K=10$ extraction rate (pass@10) by insertion
frequency for each model and training condition. Decoding uses
temperature $=1.0$, top-$p=1.0$, and 10 samples per canary. Results are
mean $\pm$ standard deviation over three seeds. Greedy results are reported
in Table~\ref{tab:full-extraction}.}
\label{tab:full-extraction-stochastic}
\small
\resizebox{\linewidth}{!}{
\begin{tabular}{llcccccc}
\toprule
Model & Training & $1\times$ & $2\times$ & $5\times$ & $10\times$ & $20\times$ & $50\times$ \\
\midrule

LLaMA-3.2-1B
& matched non-DP & 0.000$\pm$0.000 & 0.004$\pm$0.003 & 0.129$\pm$0.017 & 0.608$\pm$0.047 & 0.970$\pm$0.008 & 1.000$\pm$0.000 \\
& clip-only       & 0.000$\pm$0.000 & 0.000$\pm$0.000 & 0.000$\pm$0.000 & 0.031$\pm$0.004 & 0.373$\pm$0.017 & 1.000$\pm$0.000 \\
& DP $\varepsilon=5$ & 0.000$\pm$0.000 & 0.000$\pm$0.000 & 0.000$\pm$0.000 & 0.000$\pm$0.000 & 0.000$\pm$0.000 & 0.000$\pm$0.000 \\
\midrule

Gemma-2-2B
& matched non-DP & 0.022$\pm$0.012 & 0.100$\pm$0.014 & 0.464$\pm$0.027 & 0.878$\pm$0.031 & 0.990$\pm$0.008 & 1.000$\pm$0.000 \\
& clip-only       & 0.002$\pm$0.002 & 0.004$\pm$0.003 & 0.067$\pm$0.014 & 0.417$\pm$0.018 & 0.930$\pm$0.008 & 1.000$\pm$0.000 \\
& DP $\varepsilon=5$ & 0.000$\pm$0.000 & 0.000$\pm$0.000 & 0.000$\pm$0.000 & 0.000$\pm$0.000 & 0.000$\pm$0.000 & 0.000$\pm$0.000 \\
\midrule

Gemma-3-1B
& matched non-DP & 0.000$\pm$0.000 & 0.002$\pm$0.003 & 0.004$\pm$0.003 & 0.022$\pm$0.008 & 0.087$\pm$0.021 & 0.529$\pm$0.065 \\
& clip-only       & 0.002$\pm$0.002 & 0.000$\pm$0.000 & 0.000$\pm$0.000 & 0.014$\pm$0.008 & 0.093$\pm$0.012 & 0.817$\pm$0.051 \\
& DP $\varepsilon=5$ & 0.000$\pm$0.000 & 0.000$\pm$0.000 & 0.000$\pm$0.000 & 0.000$\pm$0.000 & 0.000$\pm$0.000 & 0.000$\pm$0.000 \\
\midrule

VaultGemma-1B
& matched non-DP & 0.000$\pm$0.000 & 0.000$\pm$0.000 & 0.000$\pm$0.000 & 0.008$\pm$0.007 & 0.063$\pm$0.005 & 0.392$\pm$0.024 \\
& clip-only       & 0.000$\pm$0.000 & 0.000$\pm$0.000 & 0.000$\pm$0.000 & 0.000$\pm$0.000 & 0.007$\pm$0.005 & 0.338$\pm$0.054 \\
& DP $\varepsilon=5$ & 0.000$\pm$0.000 & 0.000$\pm$0.000 & 0.000$\pm$0.000 & 0.000$\pm$0.000 & 0.000$\pm$0.000 & 0.000$\pm$0.000 \\
\bottomrule
\end{tabular}}
\end{table}
\subsection{Greedy and stochastic extraction.}
Our primary evaluation queries every member canary using deterministic greedy generation. Because all member canaries are evaluated rather than selected using a memorization score, this provides a population-level measure of observed extraction at each training frequency. We additionally evaluate repeated stochastic generation to capture sensitive values that may be recoverable even when greedy decoding fails. For each canary, we generate $K=10$ samples using $\texttt{temperature}=1.0$, $\texttt{top\_p}=1.0$, and $\texttt{max\_new\_tokens}=30$. We report whether each canary is recovered at least once across the ten generations. A score-selected greedy audit is
also retained as a secondary analysis for comparison with targeted memorization-based auditing, but it is not used as our primary extraction measure.
\end{document}